\documentclass[10pt,twocolumn,letterpaper]{article}

\usepackage{cvpr}              

\usepackage[dvipsnames]{xcolor}

\def\paperTitle{
Make-It-Vivid: Dressing Your Animatable 
Biped Cartoon Characters from Text
}

\def\modelname{Make-It-Vivid}

\definecolor{cvprblue}{rgb}{0.21,0.49,0.74}
\usepackage[pagebackref,breaklinks,colorlinks,citecolor=cvprblue]{hyperref}

\def\paperID{256} 
\def\confName{CVPR}
\def\confYear{2024}

\title{\paperTitle}

\author{Junshu Tang\textsuperscript{1}\footnotemark[2] \quad Yanhong Zeng\textsuperscript{2} \quad  Ke Fan\textsuperscript{1} \quad Xuheng Wang\textsuperscript{3} \\ \quad  Bo Dai\textsuperscript{2} \quad Kai Chen\textsuperscript{2}\footnotemark[3] \quad Lizhuang Ma\textsuperscript{1}\footnotemark[3] \\
\textsuperscript{1} Shanghai Jiao Tong University \quad 
\textsuperscript{2} Shanghai AI Lab \quad 
\textsuperscript{3} Tsinghua University \\
\url{https://make-it-vivid.github.io/}
}

\begin{document}
\maketitle

\renewcommand{\thefootnote}{\fnsymbol{footnote}}
\footnotetext[2]{Work done when interning at Shanghai AI Lab.}
\footnotetext[3]{Corresponding authors.}

\begin{abstract}
Creating and animating 3D biped cartoon characters 
is crucial and valuable in various applications.
Compared with geometry, the diverse texture design plays an important role in making 3D biped cartoon characters vivid and charming. 
Therefore, we focus on automatic texture design for cartoon characters based on input instructions.
This is challenging for domain-specific requirements and a lack of high-quality data. 
To address this challenge, we propose \textbf{\textit{Make-It-Vivid}}, the first attempt to enable high-quality texture generation from text in UV space. We prepare a detailed text-texture paired data for 3D characters by using vision-question-answering agents. 
Then we customize a pretrained text-to-image model to generate texture map with template structure while preserving the natural 2D image knowledge. Furthermore, to enhance fine-grained details, we propose a novel adversarial learning scheme to shorten the domain gap between original dataset and realistic texture domain. 
Extensive experiments show that our approach outperforms current texture generation methods, resulting in efficient character texturing and faithful generation with prompts. Besides, we showcase various applications such as out of domain generation and texture stylization. We also provide an efficient generation system for automatic text-guided textured character generation and animation.  
\end{abstract}    
\section{Introduction}
\label{sec:intro}


3D biped cartoon characters~\cite{luo2023rabit} breathe life into fictional characters, conveying actions, and storytelling elements engagingly. These characters find applications in various domains, including video games (\eg, Animal Crossing \cite{animalcrossing}), movies (\eg, Zootopia \cite{zootopia}), and the upcoming metaverse.
Yet, the creation and animation of these characters heavily rely on skilled artists utilizing specialized software, making it a labor-intensive and time-consuming process. 




\begin{figure}[t]
    \centering
    \includegraphics[width=\linewidth]{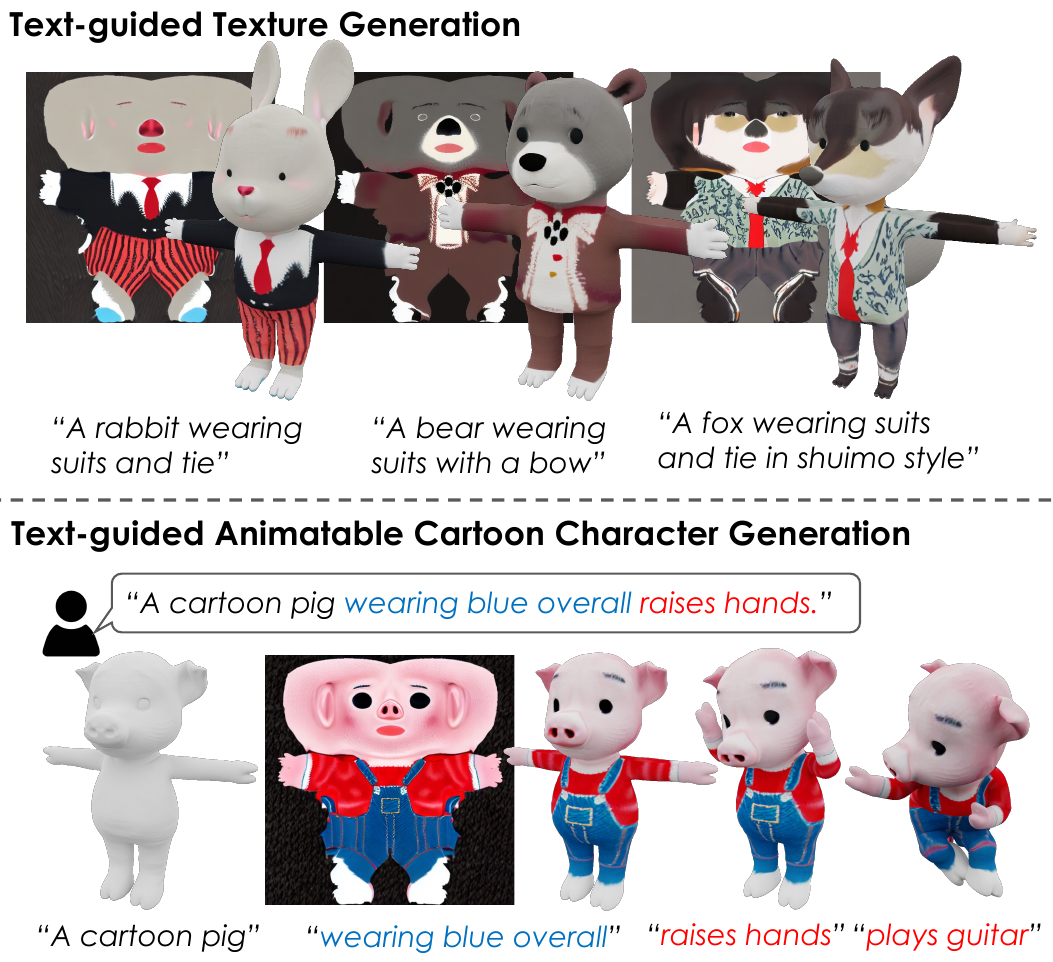}
    \vspace{-5mm}
    \caption{We present \textit{Make-it-Vivid}, the first attempt that can create plausible and consistent texture in UV space for 3D biped cartoon characters from text input within few seconds. \textit{Make-it-vivid} enables texture generation with fine-grained details in multiple styles (see above), and also supports efficient text-guided animatable textured character production (see bottom). }
    \label{fig:teaser}
    \vspace{-5mm}
\end{figure}

Compared to the shape of 3D biped cartoon characters, their textures exhibit a significantly higher level of diversity, playing a crucial role in creating vivid and charming characters. This work focuses on the automatic design of high-quality textured characters by generating textures based on text descriptions, which presents two main challenges.
\textbf{(1) Demanding domain-specific requirements.} Simply dressing 3D biped cartoon characters with appropriate textures is insufficient to make them attractive. These characters require textures that possess unique traits, including semantic harmony, consistent global configuration, and rich local high-frequency details. Consequently, conventional shape texturing methods are inadequate for cartoon characters, often resulting in textures with smooth and blurry details, as well as noticeable seam artifacts \cite{latent-nerf, fantasia3d, xmesh, tango, texture, text2tex, texfusion}.
\textbf{(2) Limited availability of high-quality data.} The scarcity of high-quality data that meets the demanding requirements further complicates the task. The creation of high-quality cartoon characters involves a costly and skill-intensive process, resulting in limited availability of such data. Additionally, due to intellectual property (IP) concerns, these data are often kept private, making it impractical to gather them from publicly accessible sources on the Internet. Existing datasets \cite{luo2023rabit} that include character texture data also suffer from significant limitations in terms of high-frequency details, inter-instance variations, and paired text descriptions.
In this work, we introduce \textbf{\modelname}, a novel texture generation framework specifically designed for 3D biped cartoon characters. Our framework enables the generation of diverse, high-fidelity, and visually compelling textures in a single forward pass, given text input.
To address the challenge of limited high-quality data, we propose marrying a knowledgeable pre-trained text-to-image (T2I) diffusion model with a topology-aware representation of the UV space, making \modelname\ the first framework to leverage diffusion priors in the UV space for 3D biped cartoon characters.
We start by developing a specialized multi-agent-based captioning system tailored for 3D biped characters. By utilizing vision-question-answering agents, we can easily generate high-quality descriptions of color, clothing, and character types based on rendered frontal views for the UV maps. This process results in a dataset of high-quality text-UVMap pairs.
Once the dataset is prepared, we customize the pre-trained T2I model to generate high-quality UV maps. This customization involves introducing learnable parameters and fine-tuning them on the paired text-UVMap data while keeping the T2I model fixed to retain its open-domain knowledge. This design allows our framework to seamlessly integrate with various customized T2I style models, such as Shuimo style \cite{moxin} and American comics, for creative texturing.

While the customized diffusion model generates various plausible textures faithful to text prompts, the texture quality often suffers from over-smoothing, making it challenging to meet demanding domain-specific requirements. This limitation can be attributed to the lack of high-frequency details in the training data. Therefore, we create high-quality images using a T2I model as a proxy for high-frequency details and innovatively introduce adversarial training \cite{goodfellow2020generative} into the diffusion training process, leading to enhanced texture details.
We extensively evaluate the performance of \modelname, demonstrating its superiority in texturing 3D biped characters. Our main contributions are as follows:
\begin{itemize} [nosep]
    \item We present \textbf{\modelname}, which empowers non-expert users to effortlessly customize vivid 3D textured characters with desired identities, styles, and attributes.
    \item To overcome the limitation of training data, we are the first to introduce adversarial training into the diffusion training process, achieving improved image fidelity.
    \item We showcase the versatility of our approach by exploring captivating applications in stylized generation and multi-modality textured character animation.
\end{itemize}

\section{Related Work}
\noindent\textbf{3D Generation under Text Guidance.}
Recent advancements in image generation~\cite{dalle,clip,stylegan2,zhang2022styleswin,stablediffusion,sdxl,imagen,deepfloydif,vqgan} have greatly boosted the research progress in 3D assets generation~\cite{dreamfields,dreamfusion,rodin,magic3d,fantasia3d,prolificdreamer} under text guidance.
A set of works~\cite{dreamfields, dreamfusion, prolificdreamer, latent-nerf, make-it-3d} propose to generate 3D shapes by optimizing a NeRF representation~\cite{mip-nerf, instant-ngp}, either through CLIP guidance~\cite{dreamfields} or Score Distillation Sampling~\cite{dreamfusion} and Variational Score Distillation~\cite{prolificdreamer} via 2D diffusion models.
Though effective, implicit representations such as NeRF can be infeasible to be deployed for most practical applications~\cite{textmesh, 3dgen-survey}.
Subsequent methods~\cite{magic3d, textmesh, clip-mesh} tackle the above problem by directly generating highly realistic 3D meshes from textual prompts.
Specifically, Magic3D~\cite{magic3d} presents a two-stage optimization framework to address the efficiency and resolution problems observed in NeRF-based models, while TextMesh~\cite{textmesh} employs an SDF backbone to extract realistic-looking 3D meshes.
There are also a number of works trying to address the different aspects of 3D content generation.
For example, Fantasia3D~\cite{fantasia3d} utilizes a hybrid representation of 3D shape, namely DMTet~\cite{dmtet}, and decouples the problem to geometry and appearance modeling.
Point·E~\cite{point-e} presents an alternative approach to 3D content generation by utilizing a point cloud diffusion model.
Latent-NeRF~\cite{latent-nerf} optimizes SDS loss in Stable Diffusion's latent space to allow increased control over the generation process.
Besides, ~\cite{latent-nerf} also presents Latent-Paint, which optimizes neural texture maps based on the input mesh.

\noindent\textbf{3D Mesh Texturing under Text Guidance.}
In addition to generate fully textured shapes, texture generation based on the given 3D geometry has recently gained significant popularity.
A number of works~\cite{fantasia3d, text2mesh, xmesh, tango, dreamfusion, latent-nerf} approach this task by optimizing an implicit representation of both 3D geometry and texture.
For example, Text2Mesh~\cite{text2mesh}, Tango~\cite{tango} and XMesh~\cite{xmesh} innovate 3D mesh texturing by optimizing the color and displacement for each vertex on the base mesh based on corresponding text prompt using CLIP loss~\cite{clip}.
While other methods~\cite{dreamfusion,sjc,magic3d,latent-nerf,fantasia3d,prolificdreamer} leverage SDS loss which helps create high-fidelity and realistic texture.
Another set of works introduce an iterative painting scheme to paint a given 3D model from different view points~\cite{texture,text2tex,texfusion}.
These methods synthesize multi-view textures based on observations under different viewpoints, and use depth-aware texture generation and inpainting to refine the new unpainted areas while preserving consistent texture from the partially painted area.
Albeit improving results, these works still suffer from severe inconsistencies across multiple views and seam artifacts due to their inpainting nature.
Furthermore, some of generative models focus on  generating high-fidelity UV textures~\cite{fukamizu2019generation,ganfit,lee2020styleuv,lattas2020avatarme,fitme,dreamface, point-uv-diffusion} directly, which shows impressive quality in 3D face reconstruction and generation.
In this paper, we explore UV texture generation on a more challenging but crucial scenario on vivid texture generation.

\begin{figure}[t]
    \centering
    \includegraphics[width=\linewidth]{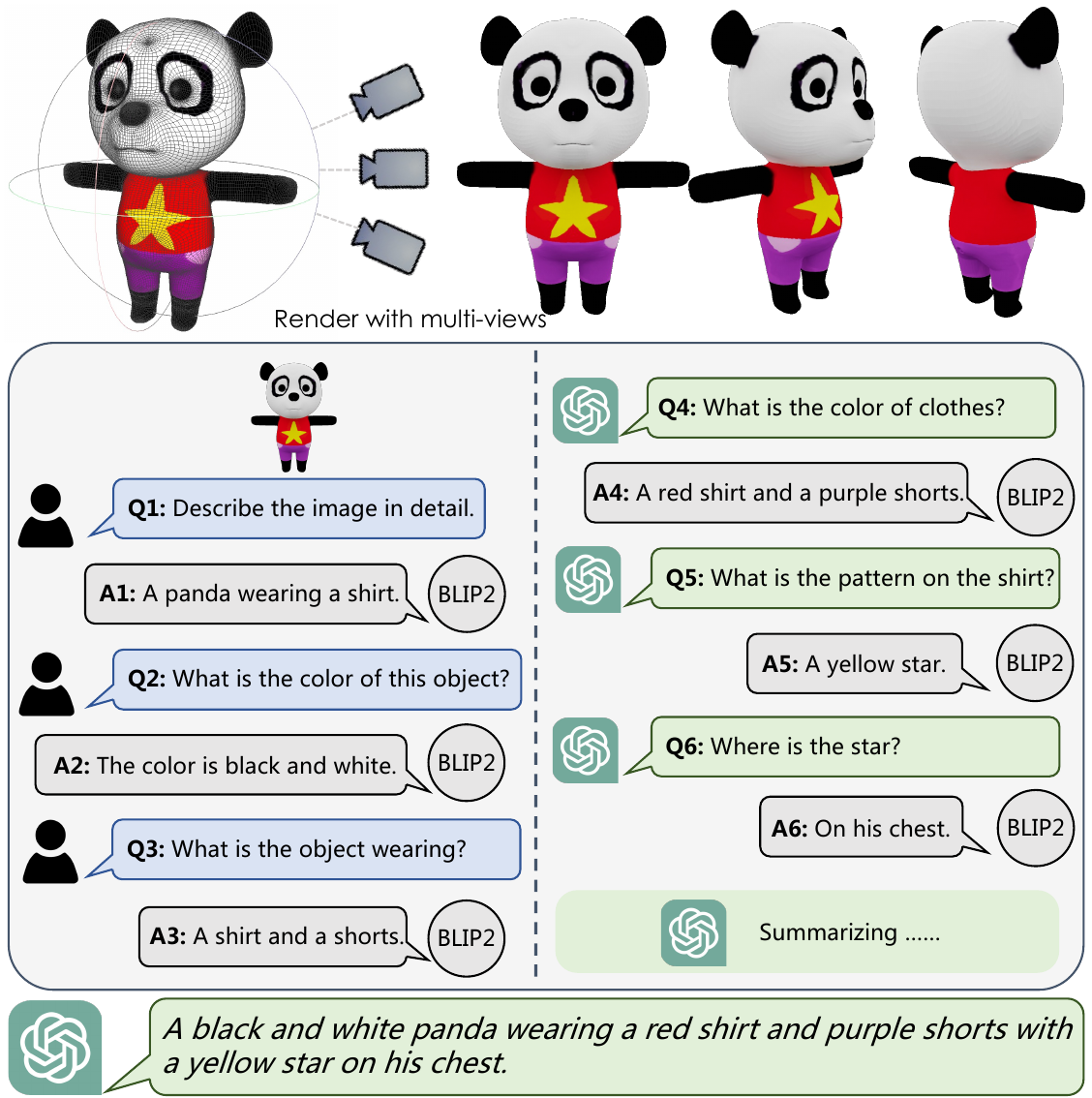}
    \caption{Multi-rounds of dialogue for captioning 3D characters. For each rendered image, we hard code three questions and use ChatGPT for asking follow-up three questions, then summarize.}
    \label{fig:chat}
\end{figure}

\section{Preliminary}

Parametric models of shapes, like 3DMM~\cite{3dmm}, FLAME~\cite{flame} and SMPL~\cite{loper2023smpl}, are widely used in computer graphics, computer vision, and other related fields. These models are designed to represent the 3D shape and appearance of complex objects, such as human bodies and faces, in a compact and expressive manner.

In order to represent 3D cartoon biped characters with a consistent topology, we adopt a parametric model Rabit~\cite{luo2023rabit} which contains a linear blend model for shapes. Specifically, the 3D biped character is parameterized as $M = F(B, \Theta, Z)$, where $B$ denotes the identity-related body parameter, $\Theta$ represents the non-rigid pose-related parameter, and $Z$ represents the latent embedding of texture.

In detail, the generated character shape is defined as $M_{S} = \bar{M}_S + \sum_{i}^{|B|}\beta_{i}s_i$,
where the mean shape is denoted as $\bar{M}_S$ and $|B|$ denotes the dimension of shape coefficients. $s_{i}\in\mathbb{R}^{3*N}$ denotes the orthogonal principal components of vertex displacements of the geometry shape. The coefficients models the variants of shapes under different identities. Besides, the shape of eyeballs can be calculated based on the predefined landmarks.

The pose parameters $\Theta = [\theta_1, \theta_2, ..., \theta_K]\in\mathbb{R}^{69}$ denotes the axis-angle of the relative rotation of joint $k$ with respect to its parents. $K=23$ denotes the number of the joints. Each parameter $\theta_k$ can be converted to the rotation matrix using Rodrigues' formular:
\begin{equation}
    \begin{aligned}
        v_i' &= \sum^{K}_{k=1}w_{k, i}G_k'(\theta, J)v_i,\\
        G_k'(\Theta, J) &= G_{k}(\Theta, J)G_k(\Theta', J)^{-1},\\
        G_k(\Theta, J) &= \prod_{j\in A(k)}[\begin{array}{cc}
           R(\theta_j) & J_{j} \\
            0 & 1
        \end{array}],
    \end{aligned}
\end{equation}
where $w_{k, j}$ denotes the skinning weight for the $i$-th vertex. $G_k(\Theta, J)$ is the global transformation of joint $k$. $J_{j}$ denotes the location of the $j$-th joint. We use this representation to animate the mesh using specific pose parameters $\Theta$.

As for parametric texture embedding, Rabit uses a StyleGAN2-based~\cite{stylegan2} generator for embedding texture map to latent codes. Specifically, the texture image $T\in\mathbb{R}^{H\times W\times 3}$ is generated by a latent code $Z\in\mathbb{R}^{d}$ where the resolution is $1024$ and the dimension $d=512$. However, it can only generate textures unconditionally with low quality. In this paper, we propose a new text-driven vivid texture generator which is editable with multiple concepts, such as color, clothes, style and so on. 





\begin{figure*}
    \centering
    \includegraphics[width=\linewidth]{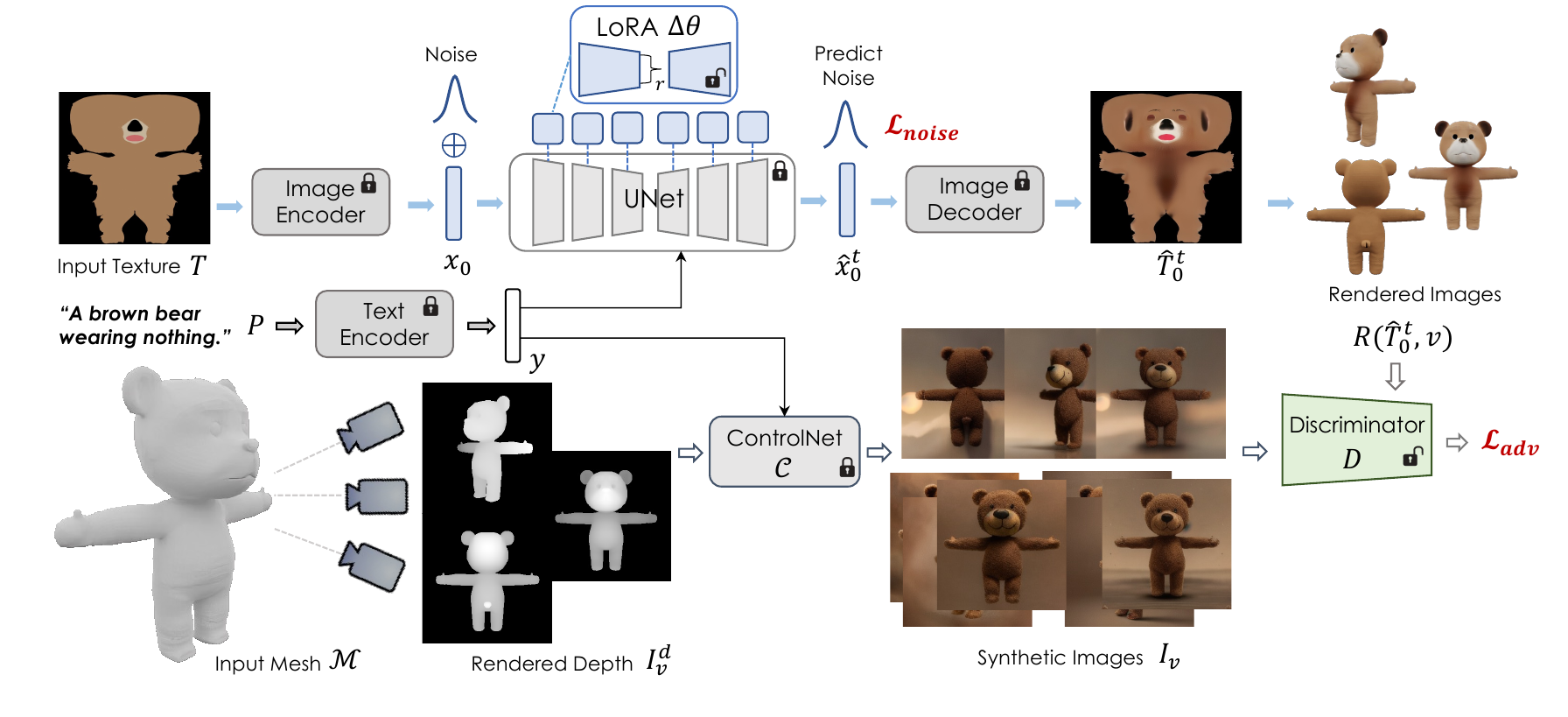}
    \vspace{-5mm}
    \caption{Overall framework for training texture generator. Our method takes a pair of data as input including a texture map $T$, corresponding text description $P$ and mesh model $\mathcal{M}$. We finetune the low-rank adaptor $\Delta\theta$ for pretrained text-to-image diffusion model to generate high quality UV texture. In order to improve the quality and perceptual fidelity of synthetic textures, we introduce adversarial training to enhance the texture details. We leverage synthetic plausible images $I_v$ conditioned by the rendered depth $I^d_v$ generated by ControlNet $\mathcal{C}$ as a proxy to guide this adversarial training.}
    \vspace{-5mm}
\end{figure*}

\section{Text-guided UV Texture Generation}
Texturing on the non-rigid cartoon character under simple instructions is crucial yet inherently challenging.
We therefore propose the first attempt to prioritize the generation of texture maps using UV unwrapping, a consistent and essential representation for mesh textures in the traditional computer graphics pipeline. 
Drawing inspiration from the achievements in image synthesis through text-conditional diffusion models, we have adopted a diffusion model for generating textures from random noise conditioned by text. 

To this end, given a user prompt $P$, our method is able to generate a vivid and consistent texture map aware of the definition of the correspondence between UV space and the geometry.
Inspired by the appealing results from the pretrained latent diffusion models, which is trained on large and diverse text-image pairs, we leverage these knowledge as semantic priors for our texture generation. To enforce the texture specifications and meanwhile preserve the generating ability, we train our texture generator by fine-tuning on the pretrained LDM.

We train our model on the 3DBiCar dataset for its topology consistent mesh model and diversity of cartoon identities. Our training samples consist of a 3D mesh, a geometry related UV texture map and a corresponding description. Since the detail description is not supported, we first propose a multi-agent character captioning pipeline for generating detail caption of each 3D textured model.

\subsection{Multi-agent Character Captioning}

In this section, we focus on building a text-texture paired data for training the texture generator.
In order to obtain the detailed description of the 3D model, inspired with~\cite{zhu2023chatgpt}, we propose a 3D caption pipeline which focuses on identity and texture information. The illustration of the caption pipeline is shown on Fig~\ref{fig:chat}.
Specifically, for each model, we firstly render multi-view images for each textured model and use Visual Question
Answering(VQA) model, BLIP2~\cite{blip}, to obtain detailed corresponding descriptions. 

Similar with~\cite{chatgptask}, we hard-code the first question as: 1) \textit{``Describe the image in detail"} to let BLIP2 generated a brief initial description of the image. Then, in order to obtain more information about the detail attribute of the color and cloth types, we start with another two specific questions to encourage the attention: 2) \textit{``What is the color of this object?"}. 3) \textit{``What is the object wearing?"}. 

Furthermore, in order to enrich image captions and generate more informative descriptions, we integrate strong vision-language model, ChatGPT~\cite{chatgpt}, for asking relevant questions according to the previous knowledge
and progressively obtain more information. 
ChatGPT is prompted to ask follow-up questions to investigate more information about the image. Besides, in order to avoid the caption model for generating pose ore action-related information, we deign a head instruction of BLIP2 including:
\textit{``Answer given questions. Don't answer any contents about the pose or action of the object."} 

At last, we use ChatGPT to summarize the descriptions across multi-views and result in the final caption. ChatGPT is able to merge the similar detail information in multi-views and remove the unlikely ones. We use the final caption as the detailed prompt of the 3D model and 3D texture for subsequent training.

\subsection{Enhanced UV Texture Generation from Text}
Now we use the prepared data for vivid and high-quality UV texture generation. 
Each pair of data includes a mesh model $\mathcal{M}$, a texture map $T\in\mathbb{R}^{H\times W\times 3}$ and a corresponding caption $P$.
To ensure the generation of similar patterns or templates of the UV map within the dataset while preserving the generating capabilities, we customize specific parameters of pretrained text-to-image diffusion model.
This customization allows us to leverage image knowledge as semantic priors in our texture generation process.
We first start with a simple baseline that a parameter-efficient finetuning, Low-Rank Adaptation (LoRA), on the U-Net of the pretrained latent diffusion model (LDM)~\cite{stablediffusion}.
We encode the input texture $T$ into latent $x_0$ and achieve diffusion process.
The objective of the training is:
\begin{equation}
    \begin{aligned}
        \mathcal{L}_{diff} = \mathbb{E}_{\epsilon, x_0, t}[||\epsilon - \epsilon_{\theta+\Delta\theta}(\sqrt{\alpha}_tx_0 + \sqrt{1-\alpha_t}\epsilon)||^{2}_{2}].
    \end{aligned}
\end{equation}
$\Delta\theta$ denotes the tuned parameters, $\epsilon$ denotes the random noise map, $\epsilon_{\theta+\Delta\theta}(\cdot)$ is the predicted noise generated by denoiser integrated with LoRA adapter, $\alpha_t$ is the parameter of noise scheduler at timestamp $t$. 

\begin{figure}[t]
    \centering
    \includegraphics[width=\linewidth]{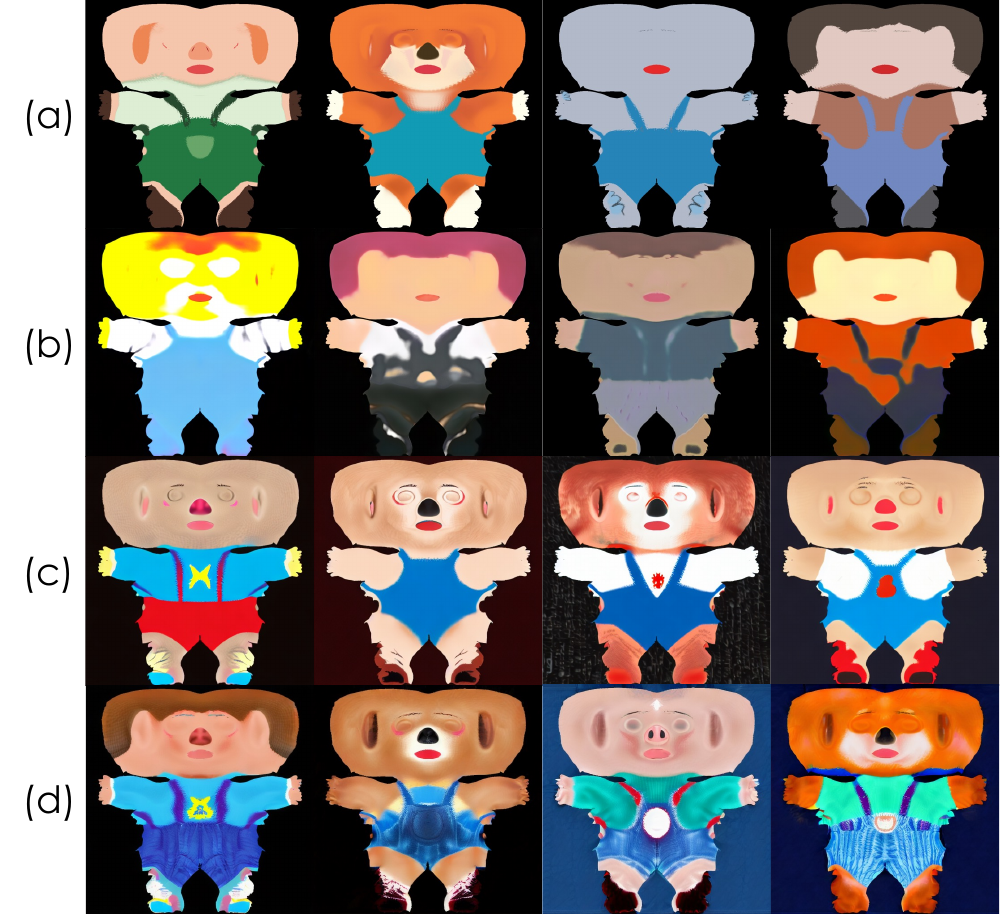}
    \vspace{-3mm}
    \caption{We showcase texture samples related with prompt \textit{``wearing overall"} (a) from 3DBiCar~\cite{luo2023rabit} dataset; (b) generated by the texture generator from Rabit~\cite{luo2023rabit}; (c) generated by the simple finetuned LDM; (d) generated by the enhanced fintuned LDM.}
    \vspace{-5mm}
    \label{fig:overall}
\end{figure}

After fine-tuning, we can infer the LDM and generate plausible texture map results related with the text instructions. 
However, simple fine-tuning on the collected dataset can only achieve UV-related texture map in the same domain with the dataset, result in limit concept and style variations.
We show some synthetic results in Fig~\ref{fig:overall} related to the text ``overall". 
We first show some selected texture maps from the dataset in (a), the selected results generated by the text generator of~\cite{luo2023rabit} in (b) and the synthetic results generated by the simple LDM in in (c). It is obvious that a simple finetuning of LDM tend to synthesis the structure of the overall without fine-grained details such as cloth wrinkles. 

Therefore, bring the original texture domain to the realistic domain is necessary as it ensures the perceptual realism of the textured 3D model. 
To fix the domain issue and boost the quality of texture with fine-grained local details, we need a set of realistic texture image to represent the realistic distribution.
However, it is hard to build or collect UV texture data that meets the requirements and related with the character geometry. Hence, we turn to create vivid synthetic images that looks like object renderings from different viewpoints.

Specifically, we apply a depth-guided image generator of ControlNet~\cite{controlnet} to produce multi-view images guided by the rendered depth.
Then we propose to impose an adversarial loss simultaneously when fine-tuning the parameters of the adapter. 
At each iteration,
we randomly sample a camera view $v$ from the pre-defined view set $\mathcal{V}$ and render the input mesh $\mathcal{M}$ to multi-view depth images. 
ControlNet receives the depth image $I^d_v$ the text prompt $P$ corresponding with the object, and, in response, synthesis high-quality images: $I_v=\mathcal{C}(I^d_v, y)$, where $y$ denotes the text embedding of $P$. In our case, we set number of renderings for each 3D characters $|\mathcal{V}|=8$.
As for generated sample, we randomly sample a timestamp $t\in (0, 1000)$ and achieve diffusion process. Then we use the pretrained decoder to
decode the denoised latent
$\hat{x}_0^t=(x_t-\sqrt{1-\alpha_t}\epsilon_{\theta+\Delta\theta}(x_t, y, t))/\sqrt{\alpha_t}$ to the image $\hat{T}_0^t$. Then we use a differentiable mesh renderer $\mathcal{R}$ to render the textured mesh with texture $\hat{T}_0^t$ at view $v$. The render output is denoted as $\mathcal{R}(\hat{T}_0^t, v)$. 
And then we adopt adversary loss to make the rendered image $\mathcal{R}(\hat{T}_0^t, v)$ has the similar local structure and perceptual realism with the generated 2D images $I_v$ at the same view.
The objective of the adversarial training can be formulated as:
\begin{equation}
    \begin{aligned}
    \mathcal{L}_{adv}=\mathbb{E}_{t,x_0,\epsilon}[logD(\mathcal{R}(\hat{T}_0^t, v)] + \mathbb{E}_{x_0}[log(1-D(I_{v}))],
    \end{aligned}
\end{equation}
where the rendered image $\mathcal{R}(\hat{T}_0^t, v)$ is considered as a fake image while the output of the ControlNet is considered as the real sample.
$D$ is the adversarial discriminator that tries to maximize $\mathcal{L}_{adv}$. 
The boost texture results is shown on Fig~\ref{fig:overall}(d). We can see that after the adversarial training, the network is able to generate more realistic texture details. 









\noindent\textbf{Texture seam fixing.}
When texturing the 3D model with synthetic UV texture image,
we find that using image generative model would inevitably ignore the consistency at the seam of the 3D model and results in black seam artifacts. This might due to the reason that each processed texture data is agnostic to the whole perspective 3D knowledge. 
To help fix this issue , we first apply a Gaussian filter around the boundary part of the texture image, which will remove the "black seam" on the back of the model. However, this cannot solve the misalignment at the boundary.
Therefore, we also conduct a simple image restoration technique on the back view of the model to mitigate the problem. 

Specifically, for the generated texture map $x_0$, we render the textured mesh using the renderer $\mathcal{R}$ to obtain the rendered image $T_v$ as seen from the back view.  
Then we apply a state-of-the-art image restoration method~\cite{diffir} to help make the rendered view perceptualy realistic without seam artifacts. Then we back project the $I_v$ to the updated texture.




\section{Experiments}

\begin{figure*}[htb!]
    \centering
    \includegraphics[width=\linewidth]{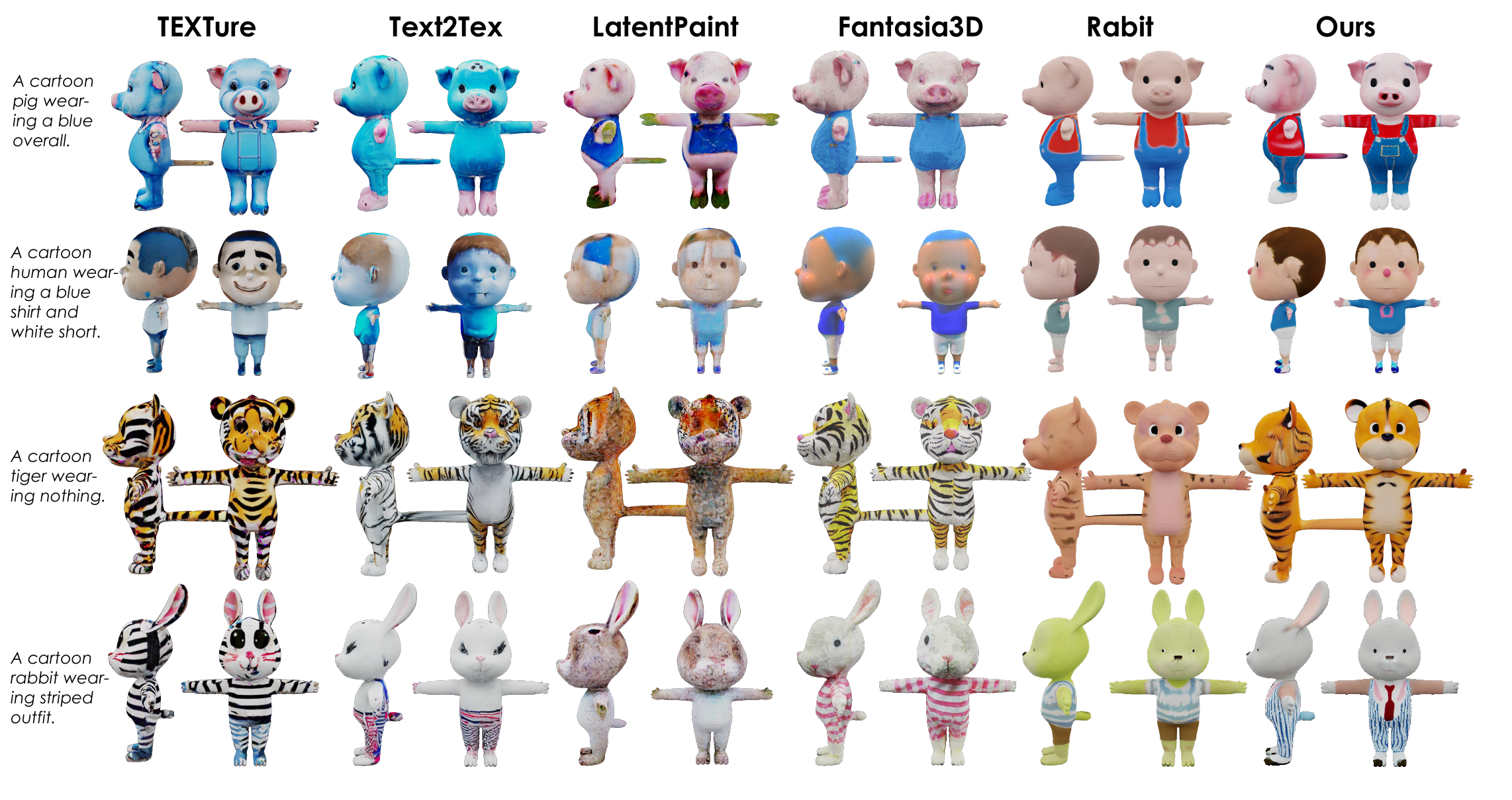}
    \caption{Qualitative comparison on the test prompt set with state-of-the-art shape texturing approaches, TEXTure~\cite{texture}, Text2Tex~\cite{text2tex}, LatentPaint~\cite{latent-nerf}, Fantasia3D~\cite{fantasia3d} and Rabit~\cite{luo2023rabit}, We show our results with high-quality and consistent texture faithful to the input prompt.}
    \label{fig:compare}
\end{figure*}

\subsection{Dataset}
Our model is trained on 3DBiCar~\cite{luo2023rabit} dataset. 3DBiCar spans a wide range of 3D biped cartoon characters, containing 1,500 high-quality 3D models. The 3D cartoon characters have diverse identities and shape resulting in 15 character species, including \textit{Human, Bear, Mouse, Cat, Tiger, Dog, Rabbit, Monkey, Elephant, Fox, Pig, Deer, Hippo, Cattle and Sheep}. All the 3D models are rigged and skinned by the predefined skeleton and skinning weight matrix, which supports further animation. Note that eyeball meshes and textures are extra modeled to support the facial expression in the future better. 
In our experiments, we use the default texture for eyeballs.

\subsection{Implementation Details}
Our texture generator is fine-tuned based on the cutting edge open source model Stable Diffusion~\cite{stablediffusion} version 1.5.
We inject LoRA into the projection matrices of query, key and value in all of the attention modules. We set the rank of the LoRA to 8. Then the modified forward pass of input $x$ is:
$h = W_0x + B_{uv}A_{uv}x$, where $B_{uv}A_{uv}$ denote the parameters of adapter. We fine-tune the adapter using the AdamW~\cite{adamw} with a learning rate $1e-4$. For inference, we use classifier-free guidance with a guidance weight $\omega: \hat{\epsilon}_{\phi}(x_{t};y,t) = (1+\omega)\epsilon_{\phi}(x_t;y,t) - \omega\epsilon_{\phi}(x_{t};t)$. In our experiments, we set $\omega=7.5$.
All the training and inference are performed on a single NVIDIA A100 GPU.

\begin{table}[t]
\centering\footnotesize
\begin{tabular}{l|ccc}
\toprule
 & CLIP$\uparrow$ & Time(min)$\downarrow$ & GPU(GB)$\downarrow$\\ \midrule
LatentPaint\protect~\cite{latent-nerf} & 27.15  & 13.95 & 11.46\\
Fantasia3D\protect~\cite{fantasia3d}  & 29.20  & 21.55 & 12.42\\
Text2Tex\protect~\cite{text2tex} & 28.81  & 14.35 & 20.31\\
TEXTure\protect~\cite{texture} &  29.25 & 2.38 & 12.05\\
\midrule
Rabit\protect~\cite{luo2023rabit} & - & 0.01 & 2.39 \\
Ours & 29.86  & 0.03 & 6.20 \\
\bottomrule
\end{tabular}
\caption{Quantitative comparison on the test prompt set with other state-of-the-art texturing method. We also report the inference time for a single prompt and GPU memory to show our efficiency.}
\label{tab:clip}
\end{table}

\begin{table}[t]
\centering\footnotesize
\begin{tabular}{l|cc}
\toprule
 & FID$\downarrow$ & KID($\times 10^{-3}$)$\downarrow$\\ \midrule
Rabit\protect~\cite{luo2023rabit} & 42.55 & 6.37 \\
Ours& 35.25& 5.25 \\
\bottomrule
\end{tabular}
\caption{Quantitative comparison on the 3DBiCar dataset. Since results of other approaches have a large domain gap with the original texture dataset, so we only compare with Rabit~\cite{luo2023rabit} trained on the same dataset.}
\label{tab:fid}
\end{table}


\subsection{Comparison with State-of-the-art Approach}
To the best of our knowledge, we are the first method focusing on texture creation in UV space of 3D biped cartoon model under text guidance. For far comparison, we build a test benchmark consisting of 300 test prompts, comprising all the 15 different species in dataset. For each species, we design 20 types of attributes about different combinations of cloth type and color.
All the prompts follow the template: \textbf{\textit{A cartoon [Species Name] wearing [Cloth Type].}}  For example: \textit{``A cartoon rabbit wearing blue shirt and white pants."}
Then we select 15 mesh models for all species from the dataset as the base mesh and texture each 3D model using corresponding text prompts. For example, we apply the texture generated from \textit{``A cartoon bear wearing suits."} on the \textit{bear} mesh. For all baselines, we render 8 views of each textured object with white background using the same renderer setting from Blender~\cite{blender} under resolution 1024*1024. 

\begin{figure}[t]
    \centering
    \includegraphics[width=\linewidth]{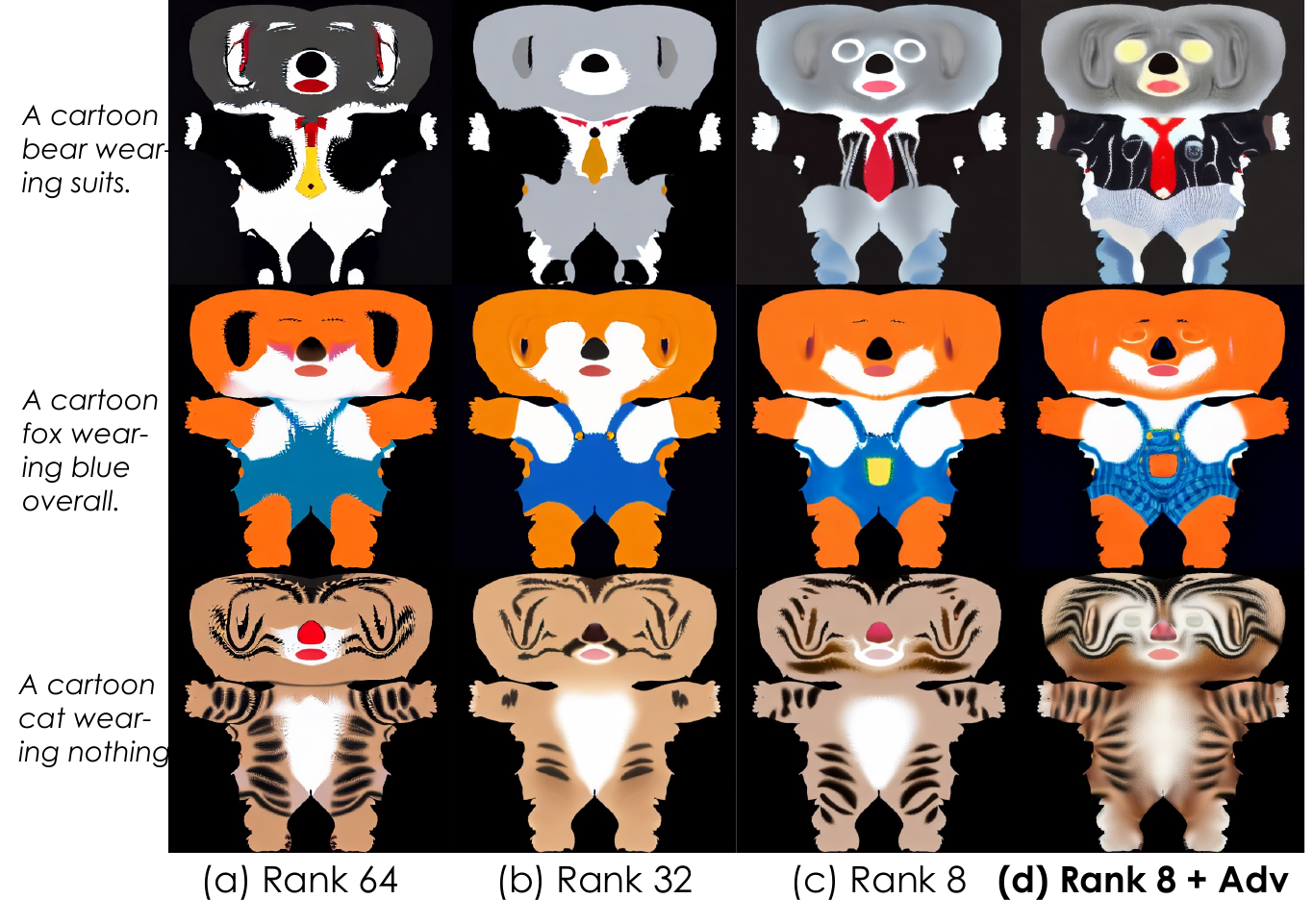}
    \vspace{-6mm}
    \caption{Ablation studies of different hyper-parameters and technical components. We visualize the synthetic results of models trained with different settings. (d) denotes our current setting.}
    \label{fig:ablation}
    \vspace{-5mm}
\end{figure}

\noindent\textbf{Baselines.}
We compare our method against two types of shape texturing approach: one is based on multi-view texture optimization, the other paints shape in a progressive manner. We first compare our approach with LatenPaint~\cite{latent-nerf} and Fantasia3D~\cite{fantasia3d}, which optimize a 3D implicit scene based on the explicit mesh guided by text under multi-view SDS loss. 
For Fantasia3D, we only initialize the DMTet~\cite{dmtet} based on the conditioned mesh model, and optimize the texture appearance under the correspondent text prompts. 
For painting methods, we compare with TEXTure~\cite{texture} and Text2tex~\cite{text2tex}, which progressively generates partial textures across viewpoints and back-projects them to the texture space. 
Besides, we also compare the results from our texture generator with the StyleGAN2-based texture generator proposed by Rabit~\cite{luo2023rabit}. 

\noindent\textbf{Qualitative comparisons.}
We compare the rendering results across several geometries textured from our approach against other baselines, as shown in Fig~\ref{fig:compare}.  
We can see that our method is able to generated consistent texture to align faithfully with the conditioned text prompt. 
In contrast, painting-based methods like TEXTure and Text2Tex have noticeable seam artifacts when viewing the side and back sides of outputs. 
Optimizing-based methods can generate multi-view consistent texture. However, LatentPaint can hardly generate high-quality and text-related neural texture. While Fantasia3D demonstrates improved rendering results, there are still noticeable non-smooth artifacts present on the surfaces.
We provide additional visualization of results using the texture generator from Rabit~\cite{luo2023rabit}. Since Rabit can only generate texture image unconditionally, so we randomly generate 100 texture maps and select relative results for visually comparison. The results show that the synthetic texture exhibits low-quality with indistinct structure. 
We conduct the user study to obtain the user's subjective evaluation of the fidelity and plausibility of the texture results. The detail can be found in \textit{Supplementary Material}.

\noindent\textbf{Quantitative comparisons.}
We evaluate the text-driven synthetic textures using average CLIP score to measure the alignment between texture image with the conditioned text prompts. The results is shown 
in Table~\ref{tab:clip}.
From the result we can see that our model achieves the best CLIP scores, indicating better text-texture alignment.
We also report the run time for generating texture under a specific text guidance using the default hyper-parameters of each method on a single GPU. Notably, our method and Rabit are significantly faster than the optimization-based methods which indicates our efficiency.
Besides, we also use the image quality and diversity metric Frechet Inception Distance (FID)~\cite{fid} and Kernel Inception Distance (KID)~\cite{kid} in Table~\ref{tab:fid}. In our experiments, on 3DBiCar dataset, the real distribution comprises renders of the geometries with the same settings using their artist designed textures. Results show that our method achieves better score than the texture generator of Rabit in terms of both FID and KID.



\subsection{Ablation Studies}

We perform extensive ablation studies on different choices of hyper-parameters and the importance of the proposed adversarial learning scheme to investigate their effects on the final results. Specifically, we vary the rank of the LoRA adapter, exploring settings of 64, 32, and 8 training without adversarial loss. Then we investigate the effect of adversarial training for texture enhancement. The visualization results are presented in Figure \ref{fig:ablation}, where the qualitative analyses unveil the influence of different settings on texture quality and diversity.
According to the visualizations, it is evident that finetuning with a large rank introduces noticeable sawtooth artifacts. While reducing the rank mitigates this issue, it concurrently leads to textures with a low-poly and excessively smooth appearance. Lower ranks, such as 8, tend to yield more plausible semantic details. Adding adversarial training will help to enhance the fine-grained patterns in the texture output. Similar visualization results are also shown on the last two rows on the Fig~\ref{fig:overall}.

\section{Applications}
\begin{figure}[t]
    \centering
    \includegraphics[width=0.95\linewidth]{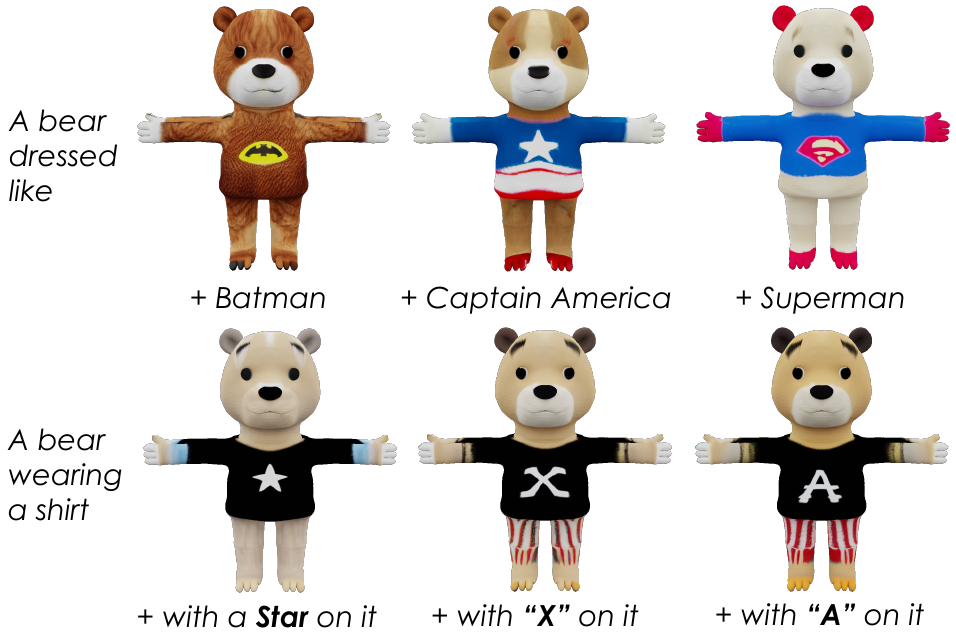}
    \caption{Make-It-Vivid enables out of domain generation about famous fictions and prompt-based local editing.}
    \label{fig:ood}
    \vspace{-3mm}
\end{figure}

\noindent\textbf{Out of domain texture generation.}
Our method could enable realistic UV texture generation that highly faithful with the text instruction, and even support out of domain generation such as fashion icons or unreal humanoid characters from famous fiction or movies while retaining high recognition. We show some of results in Fig~\ref{fig:ood}.

\noindent\textbf{Prompt-based local editing.} We also explore the controlbility of our model as a prompt-based editing method in Fig~\ref{fig:ood}. Simply using prompt-based editing can help to modify the texture according to the text while retaining other concepts. Such an editing capability makes the 3D texture creation with our model more controllable. 

\begin{figure}[t]
    \centering
    \includegraphics[width=0.95\linewidth]{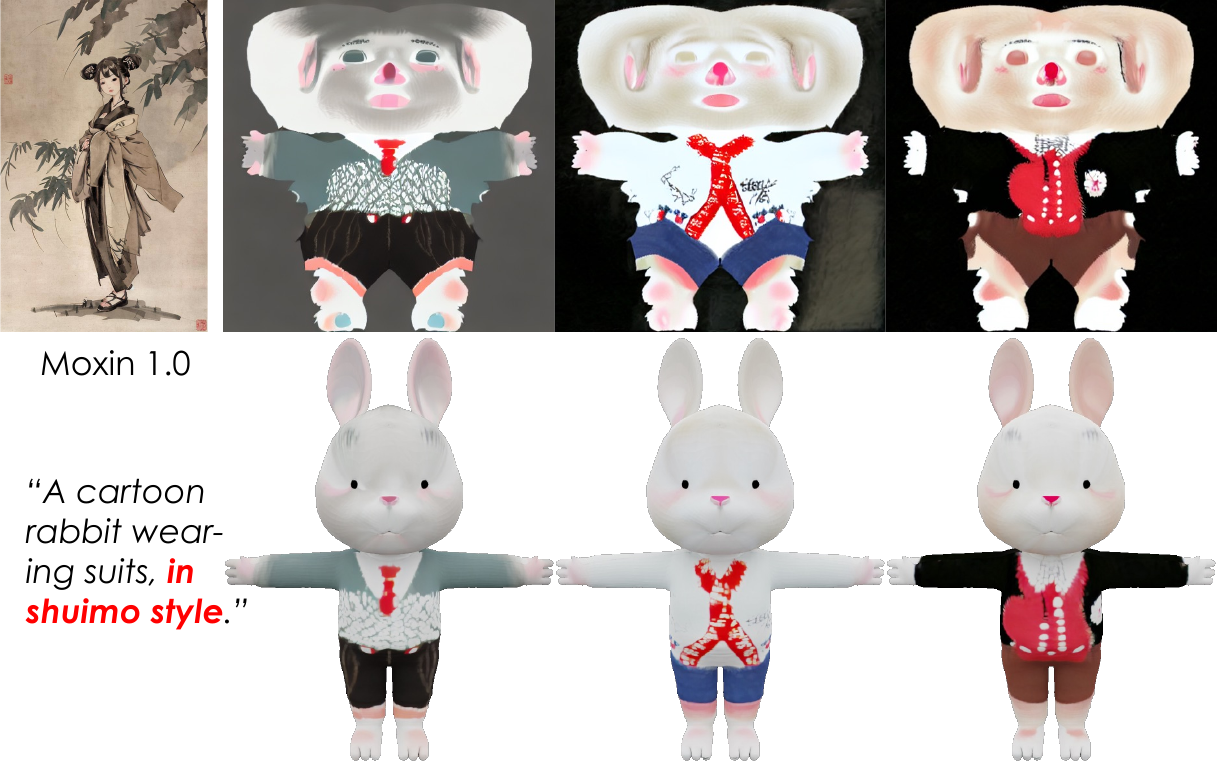}
    \vspace{-3mm}
    \caption{Make-It-Vivid enables stylized texturing. We show some synthetic results in shuimo style generated from our method injected with adapter MoXin 1.0~\cite{moxin}.}
    \label{fig:moxin}
    \vspace{-5mm}
\end{figure}

\noindent\textbf{Stylized texture generation.} Besides, we can achieve stylization for generated texture by injecting additional parameters from the other pretrained adapter $\mathcal{S}$ training on the styled image set.
Then the modified forward pass of an input x is:
$h = W_0x + B_{uv}A_{uv}x + wB_{s}A_{s}x$,
where $B_sA_s$ denotes the parameters of $\mathcal{S}$. We set the balance weight $w=0.5$.
We show some samples generated by our model and a pretrained adapter MoXin1.0~\cite{moxin} which is trained in a ink and wash painting dataset. We can see that after the stylization, the model is encouraged to generate plausible and stylized cloth types which takes large gap with original domain while preserve the original structure.

\begin{figure}[t]
    \centering
    \includegraphics[width=\linewidth]{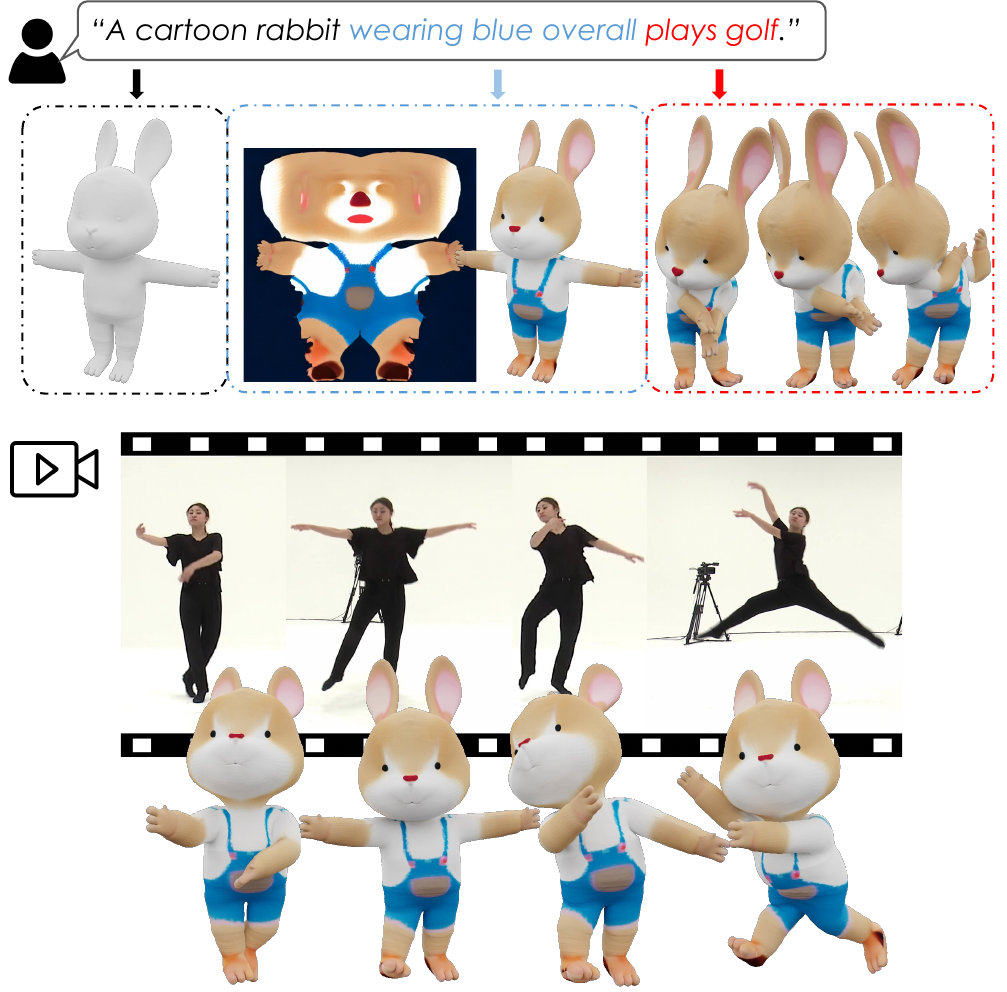}
    \vspace{-7mm}
    \caption{Make-It-Vivid supports efficient characters production and animation under text or video input.}
    \label{fig:dance}
    \vspace{-5mm}
\end{figure}

\noindent\textbf{Textured characters production and animation.}
Our method aims to help users to create and customize vivid and plausible cartoon character efficiently.
Therefore, we show the progressive generation system capable of creating textured animatable characters, driven by either text or video in Figure~\ref{fig:dance}. 
Specifically, given a text prompt, we first employ the Large Language Model (LLM)~\cite{touvron2023llama} to process the text and extract three information including subject, texture and motion.
For subject, we leverage a CLIP-based retrieval method to retrieve the shape with the nearest semantic in the dataset as the base geometry.
Then we leverage our proposed texture model to design its appearance. 
To generated related motion according to the text, we directly apply a state-of-the-art text to motion model~\cite{Guo_2022_CVPR}  to process the text and generate body rotation parameters. 
We then derive animated characters by applying the generated rotation parameters to the pre-defined joint points. 
Besides, we can also use video or other human interactions to drive or animate the created cartoon character. 


\section{Conclusion}
We propose a novel text-guided texture generation in UV space for 3D biped cartoon characters, which enables to generate high-quality and semantic plausible UV textures.
To accomplish the lack of high-fidelity data, we leverage priors from pretrained text-to-image model, which helps to generate texture map with template structure while preserving the natural knowledge. Furthermore, we propose an adversarial loss to shorten the domain gap between original dataset and realistic texture domain while training. 
Experiments show that our model can achieve efficient texture creation faithful with text input, supporting multiple stylization and local editing. Our approach can be easily applied to 3D character production and animation system, advance the 3D content creation.    
\label{sec:conclude}

\clearpage
\setcounter{page}{1}
\section*{Appendix}
\appendix

In the following, we provide additional implementation details. Later, we show the detailed quantitative ablation study and user study. Then, we discuss the limitation of our method and our future work. At last, we provide more synthetic results by Make-It-Vivid and rendered images of textured cartoon model.

\section{Additional Implementation Details}
\label{sec:imp}


\noindent\textbf{Rendering.}
During training, we implement a differentiable mesh renderer using Kaolin~\cite{KaolinLibrary} for rendering textured model. The viewpoint is randomly sampled from $\mathcal{V}: (\phi\in\{0, 45, 90, 135, 180, 225, 270, 315\}, r=1.5, \theta=80)$, where $r$ is the radius of the camera, $\phi$ is the azimuth camera angle, and $\theta$ is the camera elevation.
We composite the final rendering using generated textures with Blender~\cite{blender} rendering pipeline ``CYCLES".


%

\begin{figure}[t]
    \centering
    \includegraphics[width=0.8\linewidth]{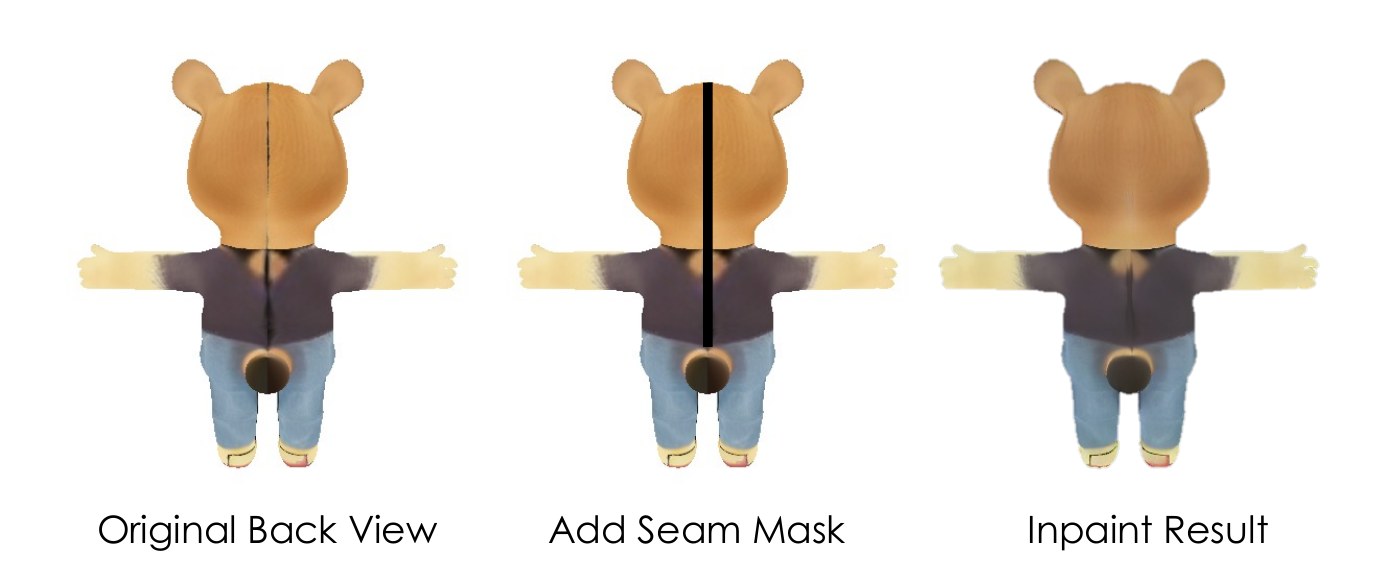}
    \caption{Illustration of texture seam fixing.}
    \label{fig:seam}
\end{figure}

\noindent\textbf{Texture Seam Fixing.} 
We conduct a simple image restoration and inpainting technique on the back view of the model to mitigate the ``black seam". 
We show the illustration of the inpainting process on Figure~\ref{fig:seam}.
Specifically, we apply a state-of-the-art image restoration method~\cite{diffir} to help make the rendered view perceptually realistic without seam artifacts. We begin by applying a line mask to the center of the image, employing an inpainting technique to seamlessly enhance the smoothness of the seam.

\section{Additional Experiments}

\subsection{Quantitative Ablation Study}
\label{sec:ablation}
We provide additional results of quantitative ablation study on different choices of hyper-parameters and the importance of the proposed adversarial learning scheme to investigate their effects on the final results in Table~\ref{tab:ablation}. We can see that decrease the rank leads to better FID and KID scores, which indicates that keep the diversity.
On the other hand, with reconstruction loss on the predicted noises only, the synthetic images contains fewer high-frequency details. In contrast, the adversarial losses can signiﬁcantly improve the image quality and perceptual realism.


\subsection{User Study}
\label{sec:user}
We conduct a user study to obtain
the user’s subject evaluation of different approaches. We present all the results produced by each comparing approaches to more than 50 participants and ask them to rank these methods on two perspectives independently: the texture quality and the fidelity with instruction. All participants are asked to rank different methods with 15 pair of comparisons in each study. 5 is for the best, and 1 for the worst.
The average score is shown in Table~\ref{tab:user}. Our method earns user preferences the best in both aspects.

\begin{table}
\centering\footnotesize
\begin{tabular}{l|cc}
\toprule
 & FID$\downarrow$ & KID($\times 10^{-3}$)$\downarrow$\\ \midrule
(a) Rank 64 & 78.52 & 21.38\\
(b) Rank 32 & 60.54 & 14.72 \\
(c) Rank 8 & 40.27 & 6.42 \\
(d) Rank 8 + Adv & \textbf{35.25}& \textbf{5.25} \\
\bottomrule
\end{tabular}
\caption{Effect of different hyper-parameters and components during training.}
\label{tab:ablation}
\vspace{-3mm}
\end{table}

\begin{table}
\centering\footnotesize
\begin{tabular}{l|cc}
\toprule
 Method & Quality$\uparrow$ & Fidelity$\uparrow$\\ \midrule
LatentPaint\protect~\cite{latent-nerf} & 1.92 & 1.45\\
Fantasia3D\protect~\cite{fantasia3d}  & 2.83  & 3.05\\
Text2Tex\protect~\cite{text2tex} &  2.72 & 2.44\\
TEXTure\protect~\cite{texture} &  2.67 & 2.16\\
\midrule
Ours& \textbf{4.53} & \textbf{4.21} \\
\bottomrule
\end{tabular}
\caption{Average ranking score of user study. For each object, users are asked to give a rank score where 5 for the best, and 1 for the worst. User prefer ours the best in both aspects.}
\label{tab:user}
\vspace{-3mm}
\end{table}

\section{Limitations and Future Work}
\label{sec:limit}
While Make-It-Vivid shows promising results on 3D biped cartoon dataset, it cannot be applied for arbitrary 3D models. That is because the collection of high-quality texture maps for all model is non-trival. Automatically generated mesh topology and UV maps are uninterpretable for 3D artists. For future work, we will put more effort on automated meshing and texturing on universial 3D data.

\section{Additional Results}
\begin{figure*}[t]
    \centering
    \includegraphics[width=\linewidth]{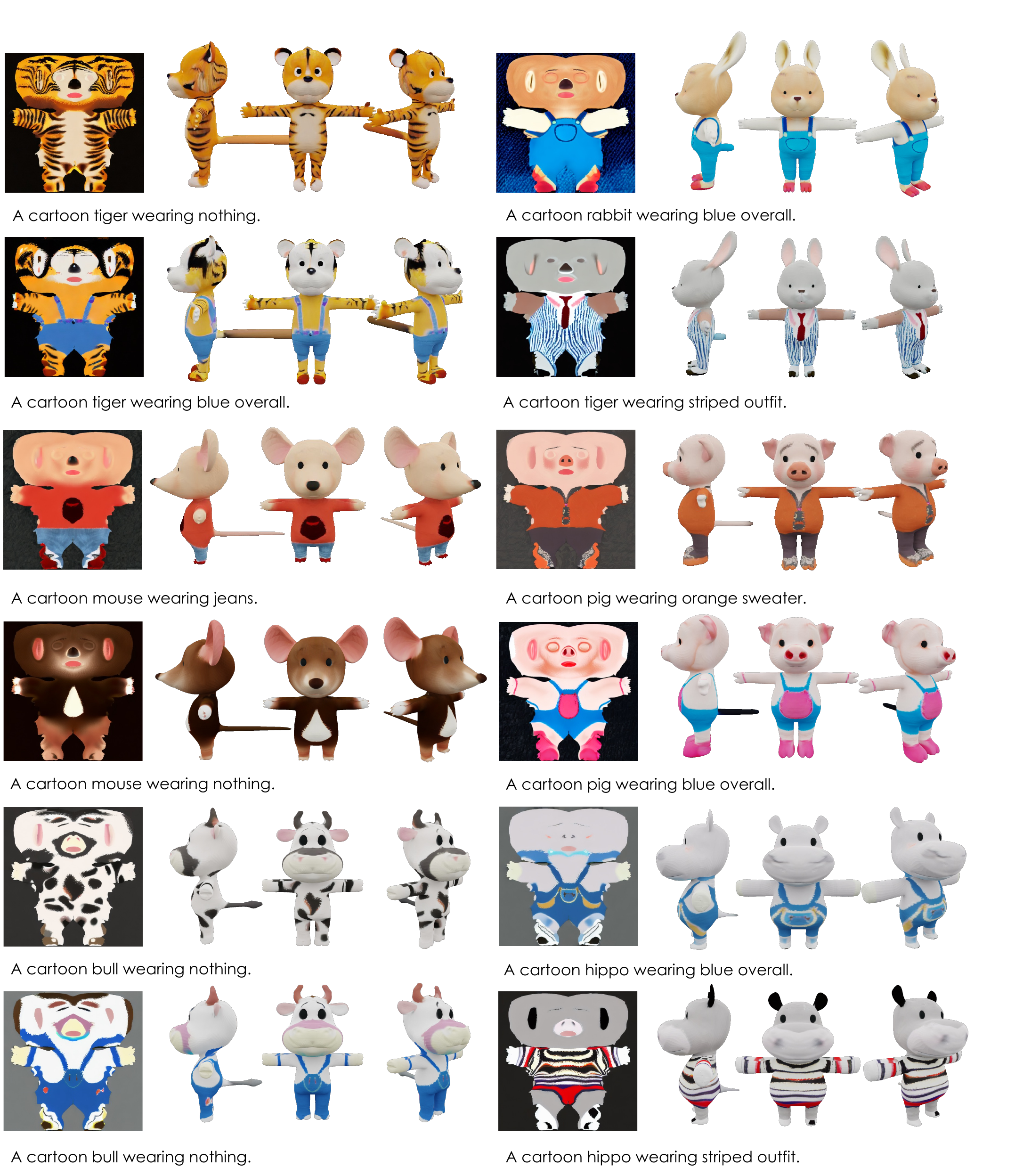}
    \caption{Additional results by \textit{Make-It-Vivid} part 1. We show the conditioned text prompt, synthetic texture map and rendered images of textured model. }
    \label{fig:more1}
\end{figure*}

\begin{figure*}[t]
    \centering
    \includegraphics[width=\linewidth]{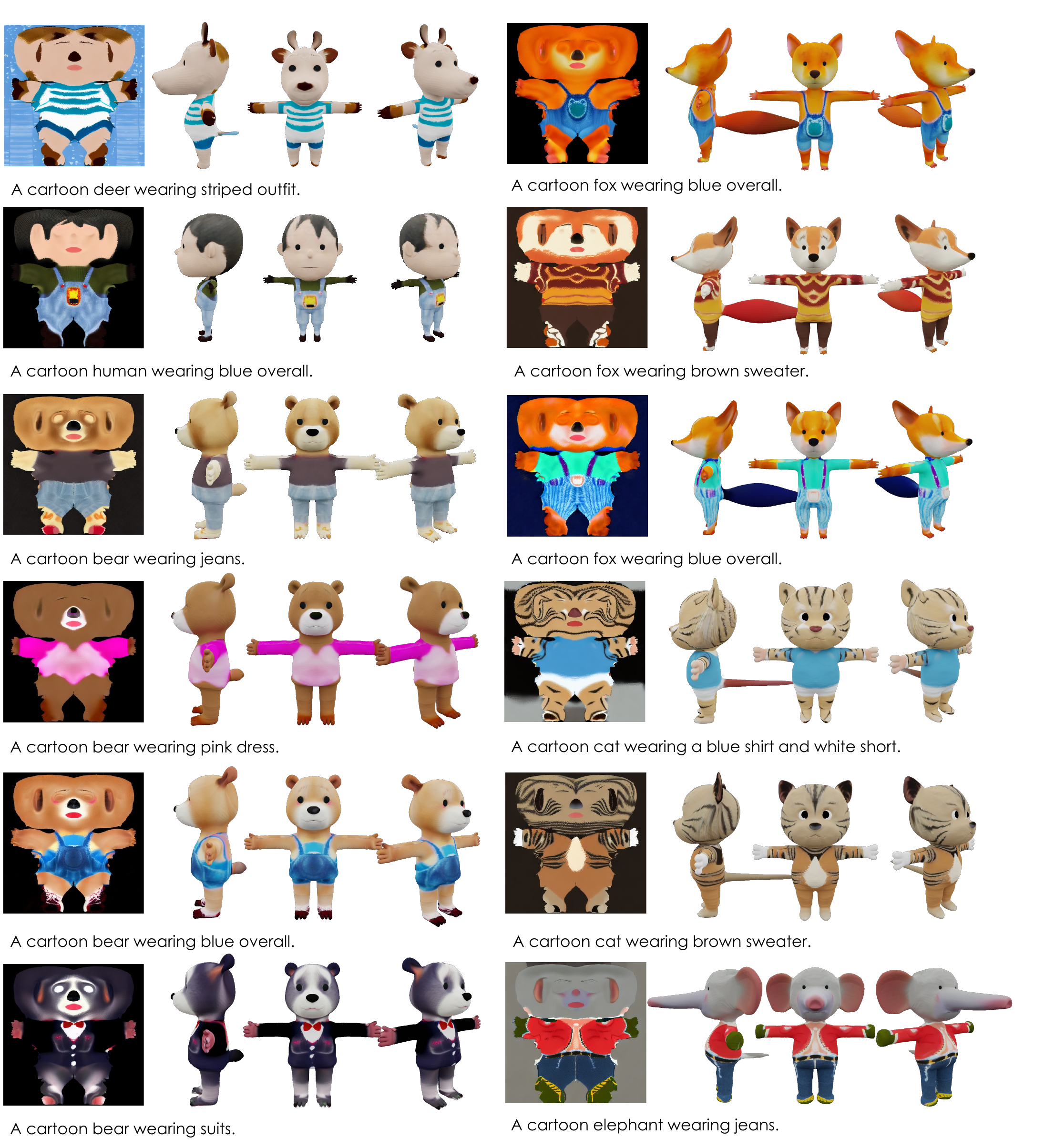}
    \caption{Additional results by \textit{Make-It-Vivid} part 2. We show the conditioned text prompt, synthetic texture map and rendered images of textured model. }
    \label{fig:more2}
\end{figure*}

In this section, we provide additional results of synthetic texture map from different text instructions using our method.
The results are shown
on Figure~\ref{fig:more1} and Figure~\ref{fig:more2}. Results show that our method has a strong ability on creating high-quality, diversity, and multi-view consistency texture map.

{
    \small
    \bibliographystyle{ieeenat_fullname}
    \bibliography{main}
}


\end{document}